\documentclass[letterpaper, 10 pt, conference]{ieeeconf}  

\IEEEoverridecommandlockouts                              

\usepackage{graphicx}
\usepackage{amsmath}
\usepackage{amssymb}
\usepackage{algorithm}          
\usepackage{algpseudocode}      
\usepackage{xcolor}

\usepackage{xcolor}
\usepackage[normalem]{ulem}

\newcommand{\redcolor}[1]{#1}
\algrenewcommand\Require{\State \textbf{Input:}}
\algrenewcommand\Ensure{\State \textbf{Output:}}
\title{\LARGE \bf
Keypoint-based Dynamic Object 6-DoF Pose Tracking via Event Camera}

\author{Zhe Wang, Qijin Song, Zihao Li, Jingyu Xiao, and Weibang Bai$^{*}$
\thanks{This work is supported by the Shanghai Pujiang Program under grant 23PJ1408500, 
and the MoE Key Laboratory of Intelligent Perception and Human-Machine Collaboration (KLIP-HuMaCo). 
The experiments in this work were supported by the Core Facility Platform of Computer Science and Communication at ShanghaiTech University. 
(*Corresponding author: Weibang Bai \textit{(Email: wbbai@shanghaitech.edu.cn)}.}
\thanks{Zhe Wang, Qijin Song, Zihao Li and Weibang Bai are with the ShanghaiTech Automation and Robotics (STAR) Center, School of Information Science and Technology, ShanghaiTech University, Shanghai, 201210, China }%
\thanks{Jingyu Xiao is with WLSA Shanghai Academy, and is an intern with the School of Information Science and Technology, ShanghaiTech University, Shanghai, 201210, China }%
}

\begin{document}

\maketitle
\thispagestyle{empty}
\pagestyle{empty}

\begin{abstract}
Accurate 6-DoF pose estimation of objects is critical for robots to perform precise manipulation tasks.
However, for dynamic object pose estimation, conventional camera-based approaches face several major challenges, such as motion blur, sensor noise, and low-light limitation. 
To address these issues, we employ event cameras, whose high dynamic range and low latency offer a promising solution. 
Furthermore, we propose a keypoint-based detection and tracking approach for dynamic object pose estimation. 
Firstly, a keypoint detection network is constructed to extract keypoints from the time surface generated by the event stream.
Subsequently, the polarity and spatial coordinates of the events are leveraged, and the event density in the vicinity of each keypoint is utilized to achieve continuous keypoint tracking.
Finally, a hash mapping is established between the 2D keypoints and the 3D model keypoints, and the EPnP algorithm is employed to estimate the 6-DoF pose.
Experimental results demonstrate that, whether in simulated or real event environments, the proposed method outperforms the event-based state-of-the-art \redcolor{methods} in terms of both accuracy and robustness.

\end{abstract}

\section{INTRODUCTION}


Object pose estimation aims to compute the precise position and orientation of a target in the world coordinate system, thereby obtaining its complete 6-DoF representation in three-dimensional space. In the field of robotics, object pose estimation plays \redcolor{a} foundational role. It provides prior pose information for autonomous grasping and placement, enabling robotic grippers to align precisely with target objects \cite{du2021vision}. For tool operation and interaction, pose estimation helps robots infer tool geometry and orientation to perform actions such as cutting, rotating, and plugging \cite{stevvsic2020learning}. In automated assembly and manufacturing, accurate pose estimation ensures precise part alignment on high-speed production lines, thereby enhancing stability and efficiency \cite{chen2020repetitive}.

In recent years, object pose estimation has made significant progress, and the types of visual sensors employed have become increasingly diverse, including monocular cameras \cite{maji2024yolo,do2018real,wang20236d}, stereo cameras \cite{pollabauer2024extending,franke20056d,ma2010probabilistic}, and depth cameras \cite{hu2021wide,kaskman2019homebreweddb}. 
%
Do et al. \cite{do2018real} introduced LieNet, a real-time framework that detects, segments, and estimates 6D object poses from a single RGB image using a rotation representation based on the Lie algebra.
Franke et al. \cite{franke20056d} proposed 6D-Vision, a Kalman filter-based stereo-motion fusion for real-time 3D position and motion estimation of image points, enabling robust obstacle detection.
These sensors have enabled a wide range of object 6-DoF pose estimation applications under static or slightly dynamic conditions \cite{do2018real, franke20056d}. 
However, the limited spatiotemporal resolution of these cameras degrades performance, making them unsuitable for accurate 6-DoF pose estimation of highly dynamic objects\cite{kim2016real}.

The event camera \cite{gallego2020event,wang2025cs3d} is a kind of bio-inspired sensor producing asynchronous events when the illumination of a single pixel changes, which makes it particularly effective in capturing detailed information of objects in high-speed motion.
Therefore, it is well-suited for 6-DoF pose estimation of highly dynamic objects.
In recent years, numerous event-based 6-DoF pose algorithms have been proposed.
%


However, existing event-based object 6-DoF pose tracking methods still exhibit significant limitations: on one hand, they show insufficient adaptability to \redcolor{non-planar geometries} \cite{liu2024line,liu2025stereo}; on the other hand, they typically rely on a predefined and aligned initial pose to enable subsequent tracking \cite{liu2024line,glover2024edopt}. 
\redcolor{However}, in service and industrial assembly environments, most components to be handled are curved-surface objects, for which existing methods show limited adaptability.
Moreover, it is often impractical to predefine and align the initial pose of the target object in real-world scenarios, thereby imposing constraints on subsequent pose tracking.

To address these challenges, we introduce a keypoint-based pipeline for object detection and pose tracking with event cameras.
The method employs keypoint detection for tracking 6-DoF pose of curved-surface objects and automatically initializes object pose via 2D-3D correspondences.
The overall pipeline is as follows:
Firstly, we introduce \redcolor{a} lightweight network for efficient multi-scale feature extraction and robust event-based keypoints detection.
Subsequently, to achieve 6-DoF pose tracking under highly dynamic conditions, this work proposes an event-based density keypoints tracking algorithm, in which temporal stability is enhanced by introducing an extended Kalman filter that effectively reduces drift and jitter.
In addition, we adopt a structure-sensitive loss that jointly enforces heatmap, coordinate, and geometric constraints, thereby ensuring accurate keypoint localization under sparse event conditions.
%

The main contributions of this work are as follows:
\begin{itemize}
\item 
\redcolor{We present a keypoint-based event-camera pipeline for 6-DoF pose tracking of dynamic objects, achieving stable and accurate estimation under high-speed motion.}
\item 
\redcolor{We propose a lightweight event-based keypoint detector that reduces target loss from delayed initial 6-DoF estimation while maintaining precise localization under sparse events.}
\item 
\redcolor{We propose an event-based density keypoint tracking method with polarity-adaptive matching and EKF for robust tracking under high-speed motion.}
\end{itemize}

\section{RELATED WORK}
Object 6-DoF pose estimation and tracking using \redcolor{different} vision sensors has been a major focus of research over the past decades. We briefly revisit the two primary categories, including those that utilize traditional RGB cameras and those that employ event cameras.

\textbf{Object Pose Tracking through Traditional RGB Cameras:} 
Recent research on RGB camera-based pose tracking applies edge extraction, feature selection, and image region processing, along with direct optimization and deep learning strategies.
Object pose estimation methods based on feature detection achieve accurate spatial pose estimation by extracting line features and keypoint features of the object. 
He et al.~\cite{he2023contourpose} propose ContourPose, a monocular 6-DoF pose estimation method that detects keypoints on object contours to establish 2D–3D correspondences for robust pose prediction of reflective, textureless metal parts.


In addition, deep learning methods have shown impressive performance in object pose estimation. 
Chen et al.~\cite{chen2022epro} propose EPro-PnP, a probabilistic and differentiable PnP solver that enables end-to-end learning of object pose from monocular images via 2D–3D correspondences. 
However, such data-driven methods struggle to achieve real-time performance, making it difficult to estimate the pose of fast-moving objects.

Due to hardware limitations, RGB cameras are prone to significant motion blur when capturing fast-moving objects, which limits their suitability for pose estimation in \redcolor{highly} dynamic scenarios.

\textbf{Object Pose Tracking through Event Cameras:}
With their high temporal resolution and low latency, event cameras offer a distinct advantage in tracking 6-DoF pose of fast-moving objects. 
The first event-camera-based algorithm for 6-DoF object pose estimation and tracking is proposed by D Reverter Valeiras~\cite{reverter2016neuromorphic}. In this work, they give object model and object's init pose information, and the algorithm estimates and tracks moving objects by associating incoming events with the 3D structure of the objects. Subsequently, Valeiras et al.~\cite{reverter2016event} present an event-based PnP algorithm to estimate object poses and enable continuous tracking, formulated as a least-squares optimization problem. Liu et al.~\cite{liu2025stereo} propose a stereo event-based 6-DoF pose tracking method for uncooperative spacecraft, leveraging geometric associations between asynchronous events and 3D object models. Without relying on deep learning, their approach employs a sparse PnP algorithm to achieve robust and low-latency pose estimation suitable for resource-constrained space applications.

Recent research in machine learning has sparked interest in utilizing data-driven algorithms for event-based object pose tracking. Rathinam et al.~\cite{rathinam2024spades} introduced SPADES, a realistic and diverse spacecraft pose estimation dataset captured using event cameras under high dynamic lighting conditions. They designed a three-channel event image encoding and occlusion mask filtering strategy, providing a high-quality benchmark for training and evaluating event-driven deep learning models while bridging the sim-to-real domain gap. Yishi et al.~\cite{yishi2025cross} proposed a 6-DoF pose estimation algorithm for non-cooperative targets by fusing monocular images with event streams. By constructing event images and employing a cross-modal Transformer architecture, their approach leverages the high dynamic range and motion robustness of event data together with the texture richness of RGB images, achieving high accuracy in challenging environments.


\begin{figure*}[t]
    \centering
    \includegraphics[width=1\linewidth]{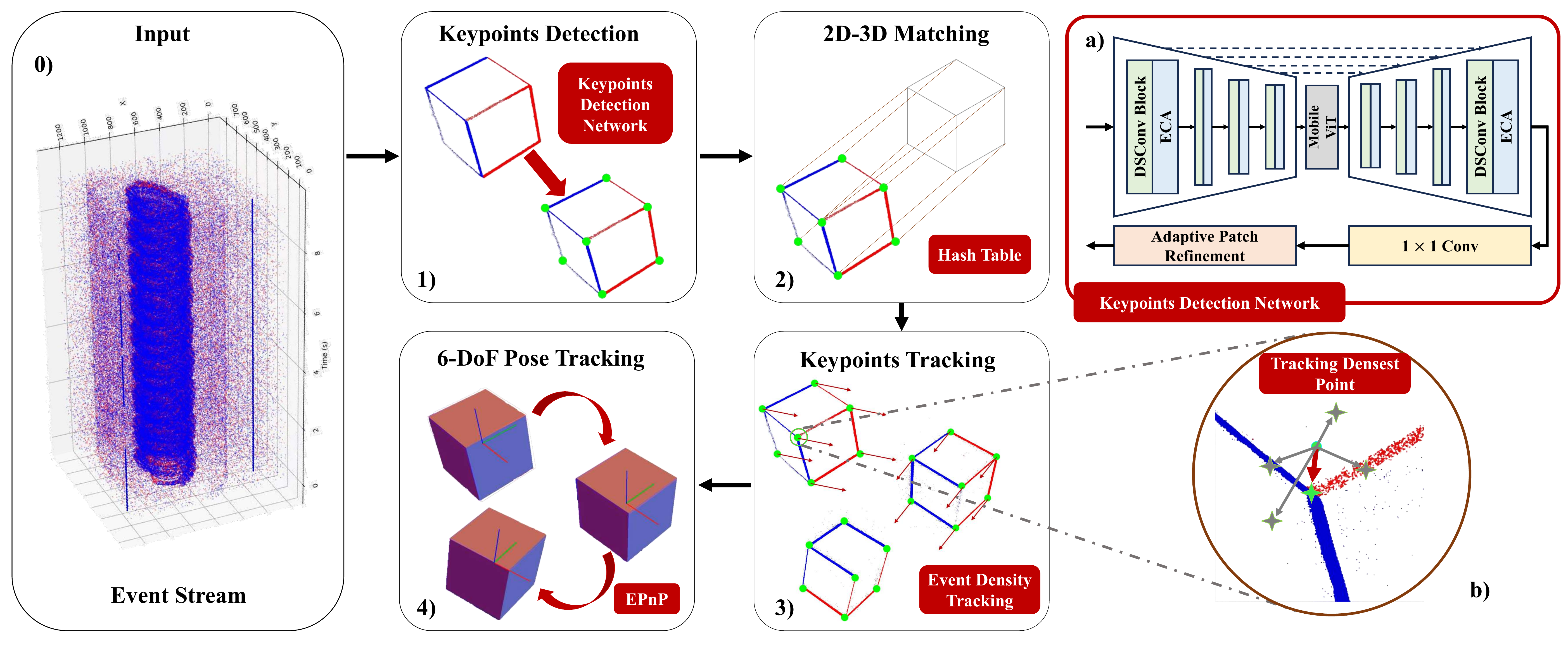}
    \caption{
    Overview of the proposed architecture. The proposed framework consists of seven building blocks: 0) obtain event stream of detection object, 1) a keypoints detection network to detect object's keypoints, 2) 2D-3D matching part using hash table, 3) event density tracking part for tracking keypoints by event density information, and 4) object 6-DoF pose tracking by EPnP \redcolor{algorithm}. a) is the architecture of this keypoints detection \redcolor{network} integrating Depthwise Separable Convolution (DSConv) \cite{howard2017mobilenets}, Efficient Channel Attention (ECA) \cite{wang2020eca}, MobileViT \cite{mehta2021mobilevit} and Adaptive Patch Refinement module. b) is a local enlarged view of step 3, illustrating the event density tracking process by highlighting the densest point used for keypoint tracking.
    }
    \vspace{-5mm}
    \label{fig:TotalPipeline}
\end{figure*}

\section{METHOD}
In this article, we propose a new event-based technique for \redcolor{tracking the 6-DoF pose of an object}.
The 6-DoF pose of the object is expressed as a homogeneous transformation matrix \( P = [R \mid T] \in \mathbb{SE}(3) \), comprising a rotation component \( R \in \mathbb{SO}(3) \) and a translation component \( T \in \mathbb{R}^3 \).

To track the 6-DoF pose of an object, we employ a four stage pipeline, as illustrated in Fig.\ref{fig:TotalPipeline}.
The first stage uses a neural network to accurately detect the keypoints of the object from the time surface \redcolor{derived from the event stream}.
The second stage matches the 2D keypoints detected by the event camera with the 3D keypoints of the model using a hash table.
The third stage leverages event information to compute the event density and integrates a Kalman filter to track keypoints.
The fourth stage involves establishing correspondences between the detected 2D keypoints and the object's 3D coordinates, followed by pose tracking via the EPnP algorithm.
\begin{figure}[t]
    \centering
    \includegraphics[width=0.80\linewidth]{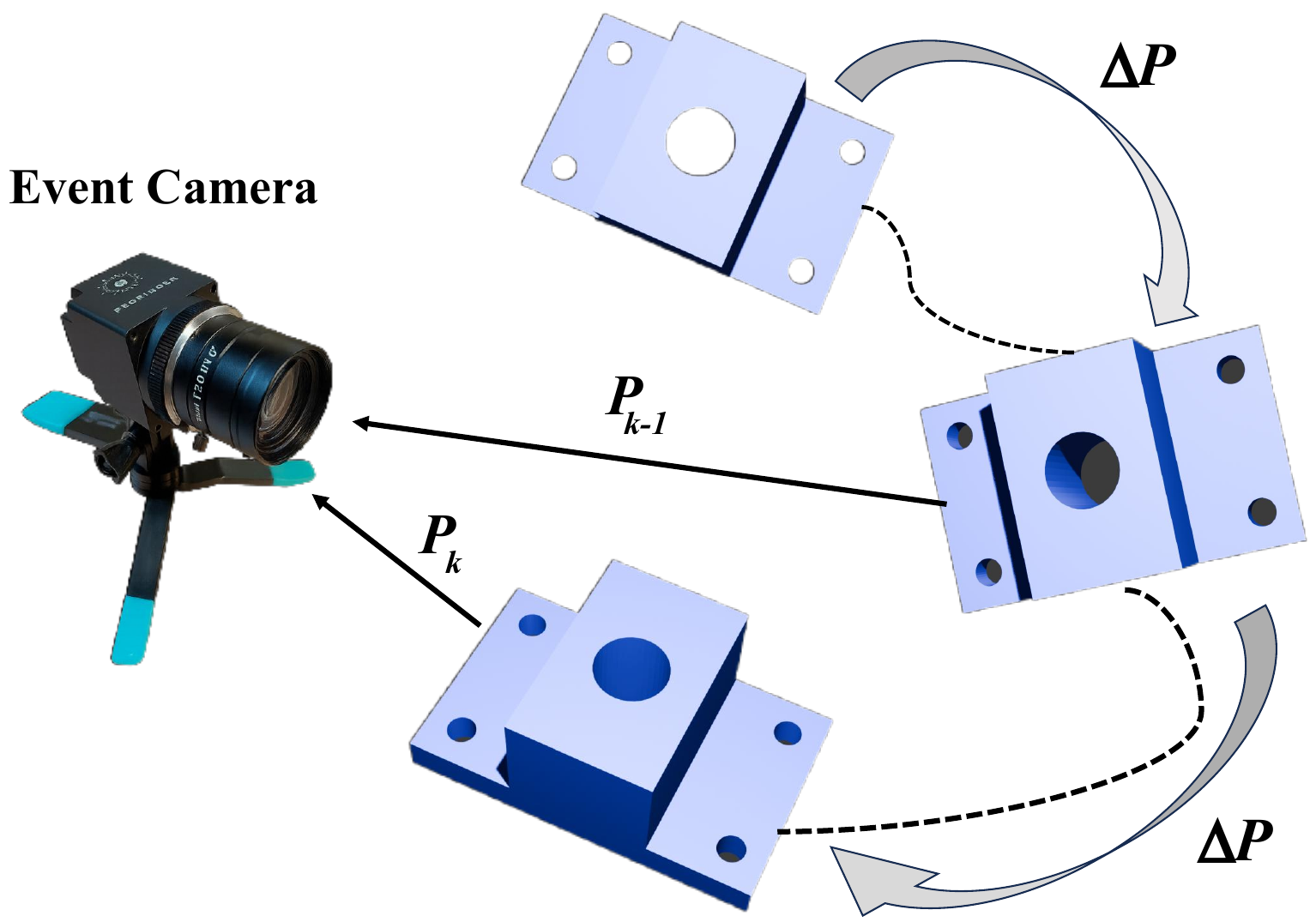}
    \caption{The geometric interpretation of 6-DoF object pose tracking. $P_{k-1}$ and $P_k$ represent the pose of the object at the previous time $t_{k-1}$ and the current time $t_k$, respectively. $\Delta P$ denotes the change in pose from $P_{k-1}$ to $P_k$.}
    \vspace{-5mm}
    \label{fig:enter-label}
\end{figure}

\subsection{Problem formulation}  
Unlike conventional cameras, which capture frames at fixed intervals, event cameras output asynchronous streams of events. 
These cameras feature \redcolor{pixels that operate independently} and respond to changes in the logarithmic photocurrent \( L \doteq \log(I) \). Events can be mathematically described as \( e_k = ( \mathbf{x}_k, t_k, p_k ) \),  where \( \mathbf{x}_k = (x_k, y_k)^\top \) denotes the location of the pixel, \( t_k \) is the timestamp of events, and \( p_k \in \{+1, -1\} \) indicates the polarity - with \( +1 \) representing a positive event (ON) and \( -1 \) indicating a negative event (OFF) \cite{chakravarthi2025recent}. 
An event is triggered when the brightness change at a single pixel exceeds a predefined threshold.
\begin{equation}
\Delta L(\mathbf{x}_k, t_k) \doteq L(\mathbf{x}_k, t_k) - L(\mathbf{x}_k, t_k - \Delta t_k)
\end{equation}

\redcolor{An event is generated when the change in logarithmic brightness at a pixel reaches a predefined contrast threshold. Specifically, the brightness change associated with an event can be expressed as}
\begin{equation}
    \Delta L(\mathbf{x}_k, t_k) = p_k D
\end{equation}

\redcolor{where $p_k \in \{+1, -1\}$ denotes the polarity of the event, and $D$ is the contrast threshold of the event camera. A positive polarity with $p_k= +1$ corresponds to an increase in brightness, whereas a negative polarity with $p_k =-1$ corresponds to a decrease in brightness.
}

In event-based pose tracking, the objective is to continuously estimate the pose \( \mathbf{P} \) of an object from the asynchronous stream of events, thereby determining its position and orientation relative to the event camera over time. 
Inspired by the concept of recursive pose estimation, the current pose of the object at time \( t_k \) can be inferred based on the previously estimated pose \( \mathbf{P}_{t_{k-1}} \) at time \( t_{k-1} \), utilizing the information provided by the incoming events at \( t_k \).
\begin{equation}
    \mathbf{P}_{t_k} = \mathbf{P}_{t_{k-1}} \, \Delta \mathbf{P}^{-1}
\end{equation}

As illustrated in Fig.~2, the geometric interpretation of 6-DoF object pose tracking indicates that the pose variation between adjacent time steps, denoted as \( \Delta \mathbf{P} \), is typically small. 
Therefore, we initialize the current pose estimation \( \mathbf{P}_{T_k} \) using the previous pose \( \mathbf{P}_{T_{k-1}} \), and iteratively refine it to obtain an accurate estimation of the current pose.

\subsection{Event-based keypoints detection} 
Because event streams provide sparse spatio-temporal information.
This makes it difficult to accurately identify the semantic keypoints of object.
To address this issue,  we design this network to enable effective extraction of semantic keypoint information, as illustrated in Fig. \ref{fig:TotalPipeline}-a.

This lightweight network integrates a UNet-based architecture with an Adaptive Patch Refinement module, thereby enabling highly accurate and robust keypoint detection.
The proposed keypoints detection network takes the time surface image of the event stream \( I \in \mathbb{R}^{H \times W} \) as input and predicts a set of keypoint heatmaps \( \{ H^k \}_{k=1}^{K} \), where each heatmap \( H^k \in \mathbb{R}^{H \times W} \) represents the probability response distribution of the \(k\)-th keypoint in the time surface image. 
The final predicted coordinate \( \hat{\mathbf{p}}^k \in \mathbb{R}^2 \) for each keypoint is obtained by extracting the peak response location from the corresponding heatmap: 
\begin{equation}
\hat{\mathbf{p}}^k = \underset{(i, j)}{\arg\max} \; H^k(i, j)
\label{eq:argmax}
\end{equation}

The overall network architecture is built upon the U-Net framework, consisting of a downsampling encoder, a semantic enhancement bottleneck, and a symmetric upsampling decoder. 
The encoder employs Depthwise Separable Convolution (DSConv) \cite{howard2017mobilenets} combined with Efficient Channel Attention (ECA) \cite{wang2020eca} to reduce computational cost. 
%

The intermediate layers incorporate the MobileViT \cite{mehta2021mobilevit} module to construct global semantic representations. 
The intermediate feature map \( F \in \mathbb{R}^{C \times H' \times W'} \) is divided into several non-overlapping patches of size \( r \times r \). 
Each patch is flattened into a token sequence \( t_i \in \mathbb{R}^{r^2 \cdot C} \), which serves as input to the transformer. 
After processing by multiple layers of the transformer encoder, the output is reshaped to its spatial dimensions and fused by convolution to reconstruct the original feature \( F' \in \mathbb{R}^{C \times H' \times W'} \). 
This module preserves local perceptual ability while establishing long-range dependencies across patches, effectively alleviating structural discontinuities in the event representation.

To further improve the accuracy and sharpness of the heatmap responses, we design an Adaptive Patch Refinement module. 
Specifically, a local patch is extracted from the coarse heatmap around the \( k \)-th keypoint:
\begin{equation}
    \mathcal{P}_k = {H}^k[x_k - r : x_k + r,\; y_k - r : y_k + r]
\end{equation}

Here, \((x_k, y_k) = \hat{p}_k\) denotes the predicted keypoint location and the patch size is \( r \times r \). 
\redcolor{The notation $[ : ]$ represents a slicing operation that extracts a local patch from the heatmap centered at $(x_k, y_k)$ with a radius $r$.}
Each patch is fed into a Tinyu-Net \cite{chen2024tinyu} module for local structural modeling and enhancement. 
The refined patch is then stitched back to the original heatmap location to improve boundary clarity and peak localization stability.

With the above architecture, the network possesses multi-scale perception capabilities from global to local levels. 
It can effectively capture both the geometric structure and response center of keypoints, which is particularly suitable for high-speed motion and sparse-event scenes in event-based vision.
Keypoints detection provides the initial position information for keypoints tracking, while continuous tracking of keypoints constitutes the necessary condition for achieving 6-DoF pose tracking.

\subsection{Loss function}
To improve both the accuracy and geometric consistency of event-based keypoints detection, we propose a Structure-Aware Keypoint Heatmap Loss, which jointly supervises the heatmap response quality, keypoint localization accuracy, and spatial structure stability. The overall loss function is formulated as a weighted combination of heatmap loss, coordinate loss, and structure preservation loss, which can be presented as:
\begin{equation}
\mathcal{L}_{\text{SAKHL}} = \lambda_1 \mathcal{L}_{\text{heatmap}} + \lambda_2 \mathcal{L}_{\text{coord}} + \lambda_3 \mathcal{L}_{\text{structure}}
\end{equation}

\noindent where \( \lambda_1, \lambda_2, \lambda_3 \) are the weights controlling the contribution of each term.

The heatmap loss function \( \mathcal{L}_{\text{heatmap}} \) measures the pixel-wise mean squared error between the predicted heatmaps \( \hat{H}_{b,k} \) and the ground-truth heatmaps \( H_{b,k} \), in each batch size $b$:
{\footnotesize 
\begin{equation}
\mathcal{L}_{\text{heatmap}} = \frac{1}{BKH W} \sum_{b=1}^{B} \sum_{k=1}^{K} \sum_{i=1}^{H} \sum_{j=1}^{W} \left( \hat{H}_{b,k}(i,j) - H_{b,k}(i,j) \right)^2
\end{equation}
}

The coordinate loss $\mathcal{L}_{\text{coord}}$ constrains the predicted keypoint locations 
$\hat{p}_{b,k}$, derived from the heatmap peaks, to remain close to the labeled ground-truth coordinates 
$p_{b,k}$ by applying the $\ell_{1}$ norm:
\begin{small}
\begin{equation}
\mathcal{L}_{\text{coord}} = \frac{1}{BK} \sum_{b=1}^{B} \sum_{k=1}^{K} \left\| \hat{p}_{b,k} - p_{b,k} \right\|_1
\end{equation}
\end{small}

The structure preservation loss function \( \mathcal{L}_{\text{structure}} \) maintains geometric consistency among all detected keypoints. It is computed by measuring the difference between pairwise Euclidean distance matrices of the predicted and ground-truth keypoint sets:
\begin{small}
\begin{equation}
\mathcal{L}_{\text{structure}} = \frac{1}{B} \sum_{b=1}^{B} \left\| D(\hat{P}_b) - D(P_b) \right\|_1
\end{equation}
\end{small}
where \( \hat{P}_b = \{\hat{p}_{b,1}, \dots, \hat{p}_{b,K}\} \), \( P_b = \{p_{b,1}, \dots, p_{b,K}\} \), and \( D(\cdot) \) denotes the pairwise Euclidean distance matrix computed from a set of keypoints.

By combining these three loss functions, the proposed structure-aware loss function enables precise, robust keypoint detection even under challenging conditions such as motion blur, sparse event data.



\subsection{Event-based density keypoints tracking}

The keypoints of an object detected by the event camera often exhibit pronounced structural saliency.
When the object is in a state of high-speed motion, these structurally salient keypoints, due to the presence of multi-directional brightness gradients at their locations, induce diverse intensity changes during motion. 
This leads to continuous triggering of both positive and negative polarity events around them, forming dense intersections of event streams. 
As a result, the event density around keypoints increases significantly. 
Therefore, to achieve robust 6-DoF pose tracking of high-speed objects, this work introduces an EKF-based event density tracking algorithm that leverages event density for matching.


In the proposed keypoints tracking method, we begin by analyzing the polarity distribution within the local region of each keypoint. 
Based on event polarity statistics, each keypoint is classified as a single-polarity point formed primarily by events of single polarity or a mixed-polarity point resulting from intersecting event lines of different polarities. 
The classification is determined by the condition described as:
%
\begin{equation}
\resizebox{0.8\hsize}{!}{$
\frac{\sum T^{+}}{\sum (T^{+} + T^{-})} > \eta \quad \text{or} \quad \frac{\sum T^{-}}{\sum (T^{+} + T^{-})} > \eta$}
\label{eq:polarity_ratio}
\end{equation}
where \( T^{+} \), \( T^{-} \) denote the total number of positive events and negative events within the local region, respectively, and \(\eta\) is the threshold parameter for classification. 
If neither condition is satisfied and the number of surrounding events is sufficiently large, the keypoint is classified as a mixed-polarity point.

For mixed-polarity keypoints, we design a polarity-aware and event-density-guided sliding window matching strategy. Specifically, within a local window centered at the \( i \)-th keypoint, for each candidate position \( (x, y) \), we extract local responses from the positive and negative polarity event time surfaces and compute the matching score as:
\begin{equation}
S_i(x, y) = \sqrt{T_i^+(x, y) \cdot T_i^-(x', y')} + \beta \cdot D_i(x, y)
\label{eq:score_function}
\end{equation}
where \( T_i^+(x, y) \) and \( T_i^-(x', y') \) denote the responses at the \( i \)-th keypoint's local patch on the positive and negative polarity event time surfaces, respectively. \( D_i(x, y) \) is the event density response in the corresponding local region. The parameter \( \beta \) is a weighting factor that enhances the contribution of structurally informative regions. Finally, the candidate location with the highest score is selected as the updated position of the \( i \)-th keypoint:
\begin{equation}
(x_i^*, y_i^*) = \arg\max_{(x, y)} S_i(x, y)
\label{eq:argmax_match}
\end{equation}

For single-polarity keypoints, when event polarity is highly imbalanced, the mixed-polarity scoring may degrade due to missing information. To address this, we design a sliding search method based solely on the single-polarity time surface, using event density response as the matching criterion.

We define the event response within the local region of a keypoint as:
\begin{equation}
D(x, y) = T^\sigma(x, y), \quad \sigma \in \{+, -\}
\label{eq:single_polarity_response}
\end{equation}
where \( T^\sigma(x, y) \) denotes the response at position \( (x, y) \) on the time surface of polarity \( \sigma \). The polarity \( \sigma \) is selected based on the dominant event type in the region surrounding the keypoint.

Within the local search region \( \mathcal{R}_i \) centered at the keypoint, we perform a dual-scale density-guided search to enhance robustness and structural discrimination. For each candidate position \( (x, y) \in \mathcal{R}_i \), we first compute the integrated event response over a large window:
\begin{equation}
S(x, y) = \sum_{(u, v) \in \mathcal{W}_b(x, y)} D(u, v)
\label{eq:density_integration}
\end{equation}

To ensure that the selected position exhibits prominent local structural features, we apply a local maximum constraint on the event response:
\begin{equation}
D(x, y) = \max_{(u,v) \in \mathcal{W}_s(x, y)} D(u, v)
\label{eq:local_max_constraint}
\end{equation}
The \( \mathcal{W}_b(x, y) \) and \( \mathcal{W}_s(x, y) \) denote the large and small window neighborhoods centered at position \( (x, y) \), respectively. 
\redcolor{The large window $\mathcal{W}_b(x,y)$ captures integrated event responses over a wider spatial region to enhance robustness against noise and event sparsity, 
and the small window $\mathcal{W}_s(x,y)$ restricts the response to a local neighborhood, ensuring that the selected position corresponds to a local maximum.}
Among all candidate positions that satisfy the above constraint, we select the one with the highest integrated event response as the updated keypoint location:
\begin{equation}
(x_i^*, y_i^*) = \operatorname*{arg\,max}_{(x,y) \in \mathcal{R}_i} \left\{ S(x, y) \;\middle|\; D(x, y)\right\}
\label{eq:final_keypoint_update}
\end{equation}

This strategy achieves robust keypoint localization by focusing on regions with the highest event density and prominent structural features. Even under conditions of high-speed motion or abrupt illumination changes that result in the loss of events of a single polarity, the method maintains accurate and stable keypoint localization.


To enhance the temporal robustness of keypoint tracking, we introduce the Extended Kalman Filter. The position and velocity of each keypoint are modeled as a state vector, which is propagated using a state transition model and updated with the observation matching the current frame. The overall estimation process can be described as:
\begin{equation}
\mathbf{x}_t = \mathbf{x}_t^{\text{pred}} + \mathbf{K}_t \left( \mathbf{z}_t - \mathbf{H} \mathbf{x}_t^{\text{pred}} \right)
\label{eq:ekf_update}
\end{equation}
where \( \mathbf{x}_t \) denotes the estimated state of the keypoint at time \( t \), and \( \mathbf{x}_t^{\text{pred}} \) is the predicted state based on the previous frame. \( \mathbf{z}_t \) represents the observed keypoint position from the current matching, \( \mathbf{H} \) is the observation matrix, and \( \mathbf{K}_t \) is the Kalman gain. This mechanism allows the keypoint to be guided by reliable matching results when available, and to rely on motion prediction when observations are missing or noisy, thereby effectively suppressing drift and jitter, and improving tracking stability. 
The overall density keypoint tracking with EKF method is illustrated in Algorithm \ref{alg:kalman_tracking}.


\begin{algorithm}[t]
    \small
    \caption{Density Keypoint Tracking with EKF}
    \label{alg:kalman_tracking}
    \begin{algorithmic}[1]
        \State \textbf{Input:} Keypoints $X$, Local time surfaces $S$
        \State \textbf{Output:} Updated keypoints $\hat{X}$
        \For{each keypoint $x_i$ and time surface $S_j$}
            \If{$x_i$ = None or $S_j$ = None}
                \State $\hat{x}_i \gets \text{None}$; \textbf{continue}
            \EndIf
            \If{EKF$_i$ not initialized}
              \State EKF$_i.\text{init}(x_i)$
            \EndIf
            \State $(p_x,p_y) \gets \text{EKF}_i.\text{predict}()$
            \State $(T^+_2, T^-_2, x_0,y_0) \gets \text{surface\_info}(S_j)$
            \State $D \gets \text{Blur}(T^+_2 + T^-_2)$
            \State Extract patch around $x_i$ from $T^+_2$ and $T^-_2$
            \State Compute polarity ratio
            \If{is single polarity}
                \State $\text{patch} \gets T^+_2 + T^-_2$
                \State $\text{peak} \gets \text{FindPeak}(\text{patch})$
                \State $\text{best\_pt} \gets \text{peak} + (x_0, y_0)$
            \Else
                \State Initialize $\text{best\_score} \gets -\infty$
                \For{each offset $(dx, dy)$ in window}
                    \State $s \gets \sqrt{T^+_2 \cdot T^-_2} + \beta \cdot D $
                    \If{$s > \text{best\_score}$}
                        \State $\text{best\_pt} \gets (dx + x_0, dy + y_0)$
                        \State $\text{best\_score} \leftarrow s$
                    \EndIf
                \EndFor
            \EndIf
            \State EKF$_i.\text{update}(\text{best\_pt})$
            \State $x_{upd}, y_{upd} \gets \text{KF.state}$
            \State $x \gets \alpha \cdot x_i + (1-\alpha) \cdot x_{upd}$
            \State $y \gets \alpha \cdot y_i + (1-\alpha) \cdot y_{upd}$
            \If{$(x, y)$ inside window}
                \State $\hat{x}_i \gets (x, y)$
                \State Append to trajectory
            \Else
                \State $\hat{x}_i \gets \redcolor{(x_i,y_i)}$
            \EndIf
        \EndFor
        \State \Return $\hat{X}$
    \end{algorithmic}
\end{algorithm}

\section{EXPERIMENTS}

 In this section, we evaluate the performance of our proposed object 6-DoF pose tracking method on both simulated and real event datasets. The experimental protocol is structured as follows: Section~\ref{sec:ImplementationDetails} describes the detailed implementation of our method; Section~\ref{sec:EvaluationMetrics} introduces the metrics used for tracking assessment. 
 Then, we present the results of simulated experiments in Section~\ref{sec:SimulatedEventExperiment} and real scenario experiments in Section~\ref{sec:RealEventExperiment}.

\subsection{Implementation Details}
\label{sec:ImplementationDetails}

\begin{figure}[t]
    \centering
    \includegraphics[width=0.98\linewidth]{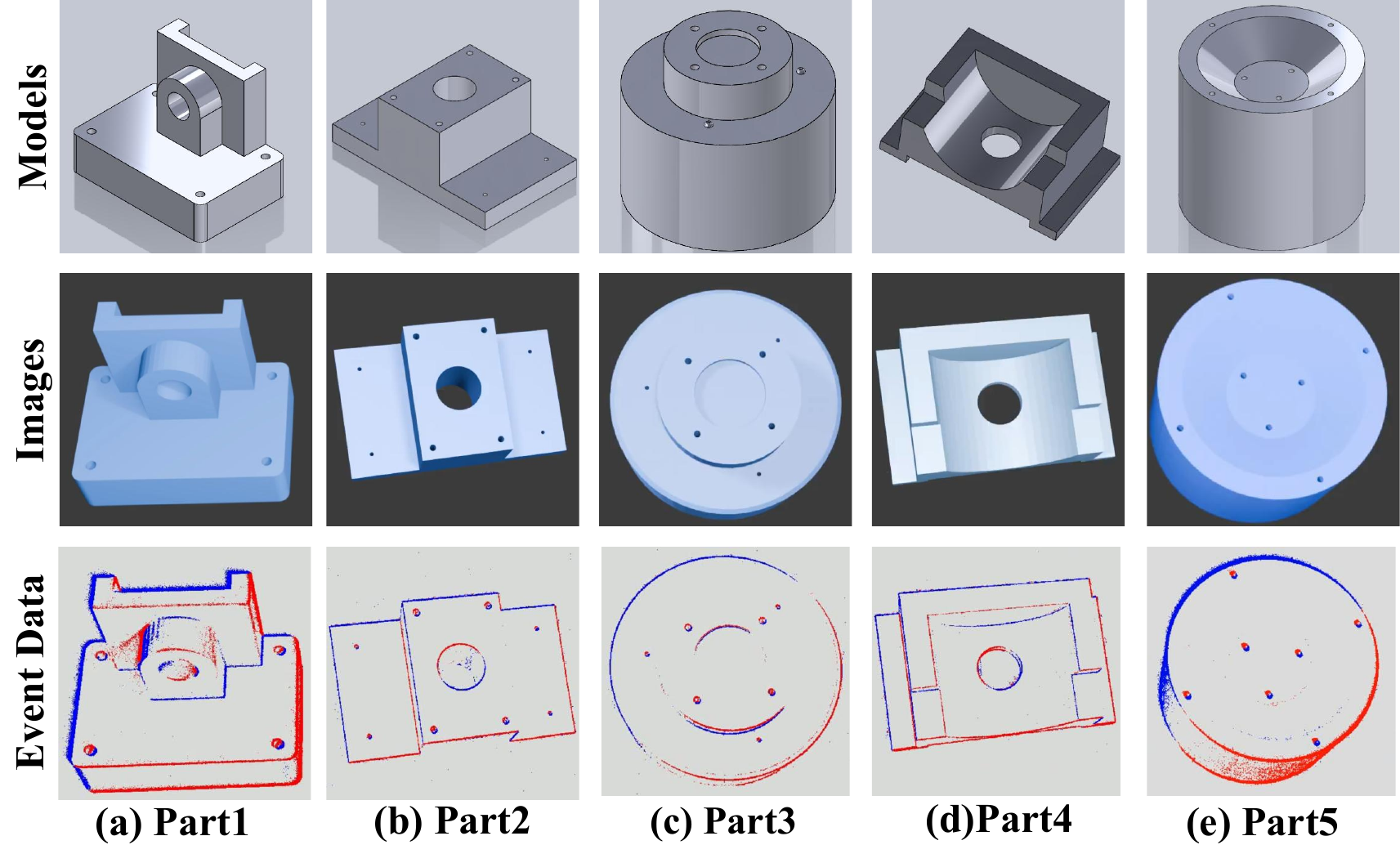}
    \vspace{-1mm}
    \caption{The models, images, and events data of mechanical parts in simulated event datasets. The top row shows the CAD models of mechanical parts, the middle row shows rendered RGB images of mechanical parts, and the bottom row displays event-accumulated images of those mechanical parts.}
    \vspace{-4mm}
    \label{fig:SimulationModel}
\end{figure}

In the simulation event experiment, we select several representative mechanical part models, their CAD models, rendered RGB images, and event-accumulated images are shown as illustrated in Fig. \ref{fig:SimulationModel}. 
The selected models include objects with prominent straight-edge structures as well as those with complex curved contours, to validate the effectiveness of the proposed method in industrial applications.

The steps of the simulation experiment are as follows. 
Firstly, we employ Blender to render RGB videos of object motion while simultaneously recording the object’s pose variations as ground truth annotations.
 Subsequently, these videos are converted into event data using the V2E tool \cite{hu2021v2e}. 
Finally, the error is quantified by comparing the pose tracked by our method with the ground-truth pose computed in Blender.





In the real event experiment, we 3D-print two models derived from the simulation dataset, including a line-based model and a curve-based model, to evaluate the accuracy of our proposed approach. Experiments are performed with an Intel(R) Core(TM) i7-13700 and an NVIDIA GeForce GTX TITAN X GPU (12GB VRAM)



\begin{figure}[h] 
    \centering
    \includegraphics[width=0.49\textwidth]{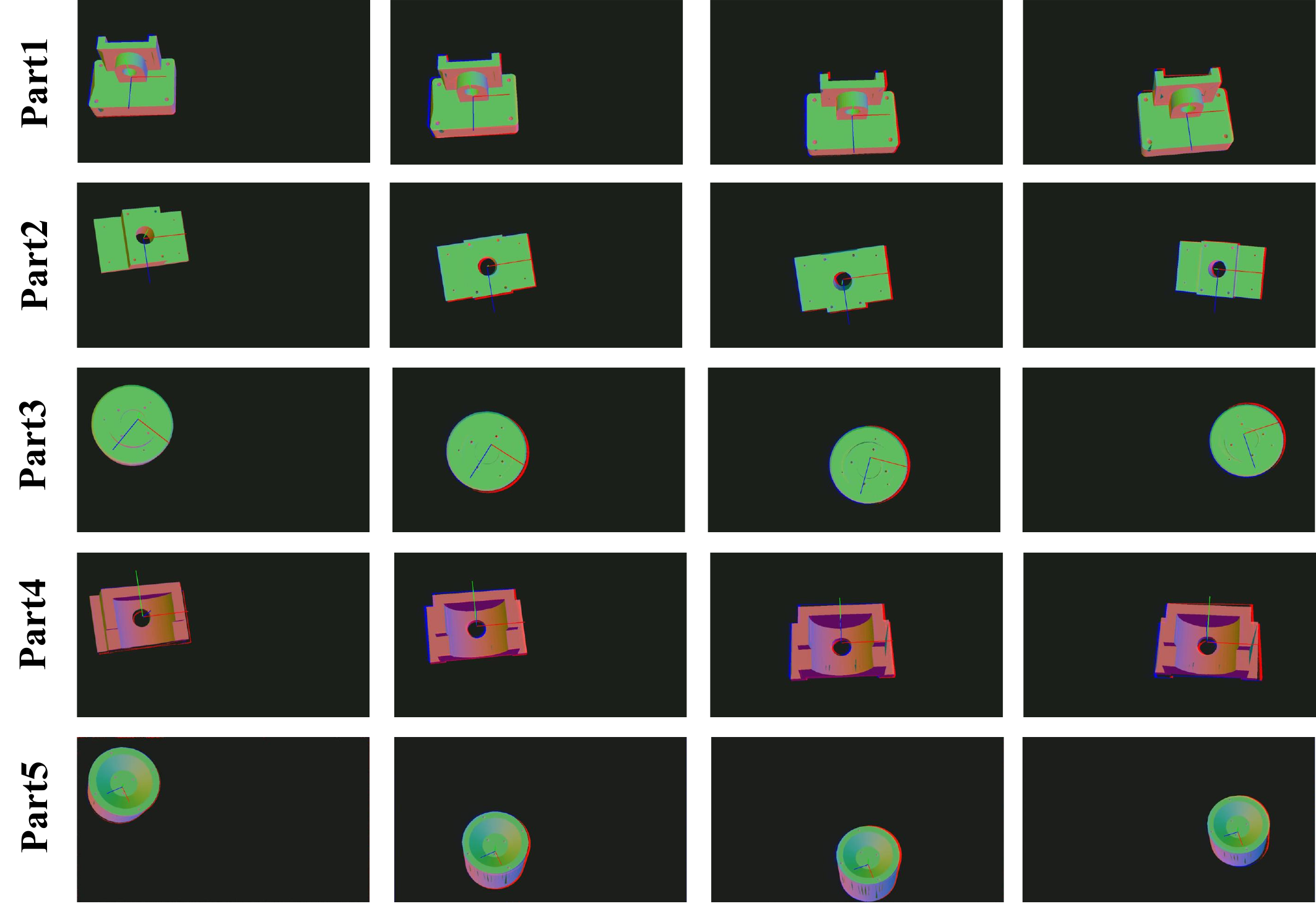}
    \vspace{-3mm}
    \caption{The simulated event experiment results of high-speed motions, including the rendered model of each object and the 3D coordinate axes representing the predicted object poses.}
    \label{fig:ObjectPose}
    \vspace{-3mm}
\end{figure}

\subsection{Evaluation Metrics}
\label{sec:EvaluationMetrics}
To assess the accuracy of the proposed method, we evaluate the 6-DoF object poses using a widely adopted protocol \cite{sturm2012benchmark} that incorporates the relative pose error.

To evaluate the accuracy of pose tracking, we adopt the relative pose error, denoted as $R_{rel}$ and $T_{rel}$, which measures the discrepancies in rotation and translation between the estimated results and the ground truth. 
It can be formulated as follows:
\begin{equation}
\mathbf{E}_{i} = \left( \mathbf{Q}_{i}^{-1} \mathbf{Q}_{i+\Delta} \right)^{-1} 
\left( \mathbf{P}_{i}^{-1} \mathbf{P}_{i+\Delta} \right)
\end{equation}
where $Q_i$ and $P_i$ denote the ground truth pose and the estimated pose at the time step $i$, respectively. 

Subsequently, the relative rotation error $R_{\text{rel}}$ and the relative translation error $T_{\text{rel}}$ are derived from the rotational and translational components of $E_i$:
\begin{align}
    \resizebox{0.85\hsize}{!}{$
R_{\text{rel}} =
\sqrt{\frac{1}{m} \sum_{i=1}^{m}
\left(
\frac{\arccos\!\left(\tfrac{\mathrm{trace}(\mathrm{Rot}(E_i)) - 1}{2}\right)}{\Delta t}
\right)^2 } \label{eq:R_rel} $} \\
\resizebox{0.60\hsize}{!}{$
T_{\text{rel}} =
\sqrt{\frac{1}{m} \sum_{i=1}^{m}
\left(
\frac{\left\| \mathrm{trans}(E_i) \right\|}{\Delta t}
\right)^2 }
\label{eq:T_rel}
$}
\end{align}


These metrics quantify the relative motion between the ground truth and the estimated trajectory over the same time interval, thereby providing a measure of the local pose drift.

\begin{figure}[b]
    \centering
    \includegraphics[width=0.99\linewidth]{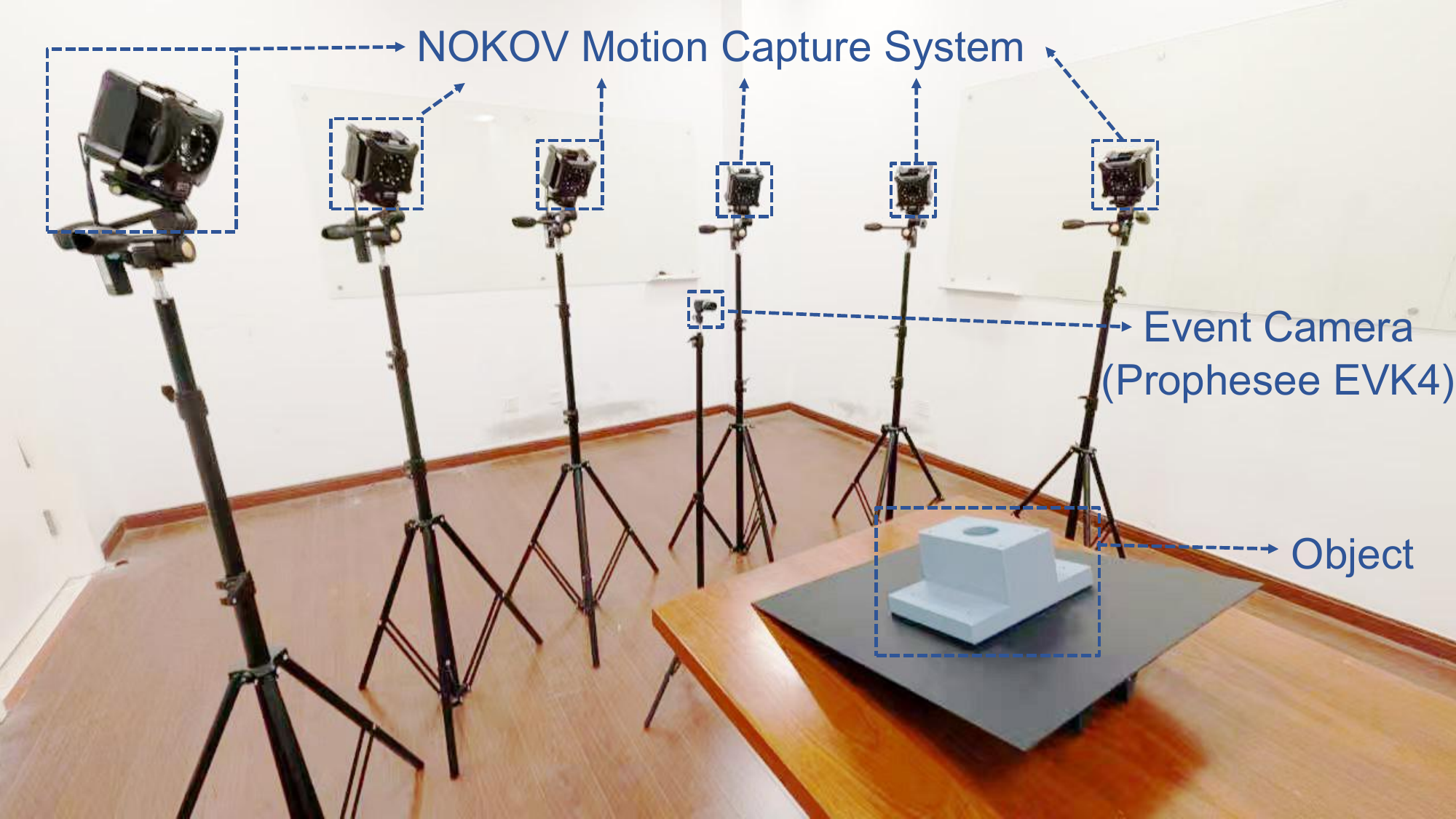}
    \caption{Experimental scenarios of real events.}
    \label{fig:SceneLayout}
\end{figure}

\subsection{Simulated Event Experiment}
\begin{figure*}[h] 
    \centering
    \includegraphics[width=0.95\textwidth]{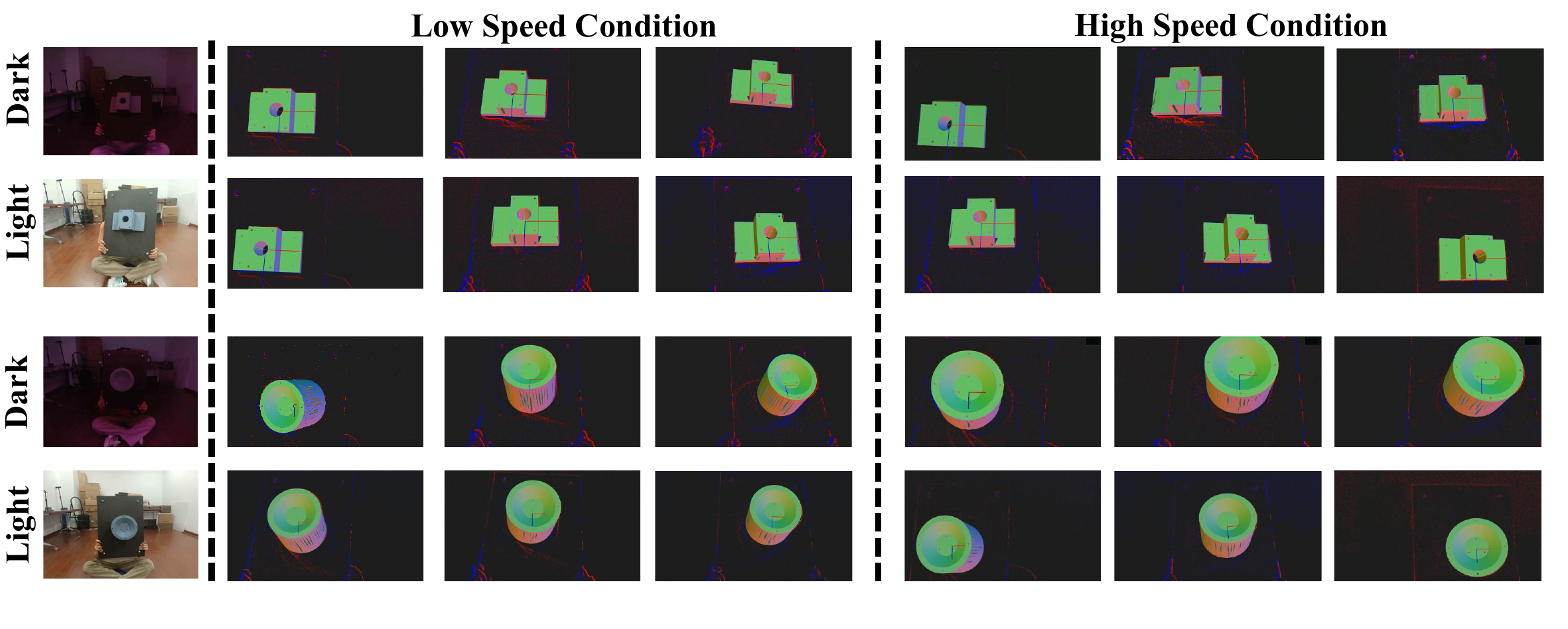}
    \vspace{-3mm}
    \caption{Pose tracking results of part2 and part5 of real event experiments, including the rendered model of each object and the 3D coordinate axes representing the predicted object poses.}
    \vspace{-2mm}
    \label{fig:RealPose}
\end{figure*}
\label{sec:SimulatedEventExperiment}
Currently, there is still a lack of event camera datasets specifically designed for mechanical parts 6-DoF pose estimation tasks. 
Although some existing works have constructed relevant datasets, these have not been publicly released, thus hindering open replication and fair comparison.
To address this limitation, we introduce an event-based dataset of irregularly moving mechanical parts, comprising both simulated and real events, to facilitate accurate evaluation of the proposed methods in object 6-DoF pose tracking.



In simulated experience, we select five types of mechanical parts with distinct linear and curved features as test objects. Each object is assigned different motion trajectories and velocities, to fully evaluate the adaptability and robustness of the algorithm under various motion patterns. 
The motion of objects is captured by a stationary monocular camera at a sampling rate of 60 Hz and a resolution of 1280 × 720 pixels, which is consistent with the resolution of the Prophesee EVK4 event camera.
Subsequently, the video captured by the monocular camera is converted into event data using the V2E tool.

\begin{table}[h]
    \centering
    \renewcommand{\arraystretch}{1.1} 
    \setlength{\tabcolsep}{3 pt}
    \caption{Comparison of relative rotation error and relative translation error on simulated event datasets [$R_{rel}$: deg/s, $T_{rel}$: cm/s].}
    \begin{tabular}{c|cc|cc|cc}
        \hline
        Method & \multicolumn{2}{c|}{Condition} & \multicolumn{2}{c|}{EDOPT\cite{glover2024edopt}} & \multicolumn{2}{c}{Ours} \\
        \hline
        Model & linear velocity& angular velocity & $R_{rel}$ & $T_{rel}$   & $R_{rel}$ & $T_{rel}$ \\
        \hline
        Part1 & $2.92 \,\text{m/s}$ & $8.27 \,\text{deg/s}$ & 0.81 & 4.91  & \textbf{0.48} & \textbf{2.97}  \\
        Part1 & $5.84 \,\text{m/s}$ & $16.54 \,\text{deg/s}$ & 0.88 & 2.38 & \textbf{0.28}  & \textbf{1.77} \\
        Part2 & $3.51 \,\text{m/s}$ & $31.64 \,\text{deg/s}$ & 1.51 & 2.00  & \textbf{0.32} & \textbf{1.91} \\
        Part2 & $7.02 \,\text{m/s}$ & $63.28 \,\text{deg/s}$ & 1.04 & 4.93  & \textbf{0.37} & \textbf{1.86} \\
        Part3 & $2.80 \,\text{m/s}$ & $48.81 \,\text{deg/s}$ & 7.55 & \textbf{3.23} & \textbf{0.69} & 4.36 \\
        Part3 & $5.59 \,\text{m/s}$ & $97.62 \,\text{deg/s}$ & 7.88 & \textbf{3.32}  & \textbf{0.43} & 4.70 \\
        Part4 & $2.47 \,\text{m/s}$ & $23.82 \,\text{deg/s}$ & 4.47 & 6.20 & \textbf{0.80} & \textbf{3.26} \\
        Part4 & $4.94 \,\text{m/s}$ & $47.64 \,\text{deg/s}$ & 5.12 & 4.22 & \textbf{0.67} & \textbf{3.71} \\
        Part5 & $4.12 \,\text{m/s}$ & $29.85 \,\text{deg/s}$ & 7.86 & 5.40  & \textbf{0.62} & \textbf{1.73} \\
        Part5 & $8.24 \,\text{m/s}$ & $59.70 \,\text{deg/s}$ & 6.71 & 5.55  & \textbf{0.63} & \textbf{1.62} \\
        \hline
    \end{tabular}
    \vspace{-2mm}
    \label{tab:Simulationcomparison}
\end{table}

The experimental results of object 6-DoF pose tracking under simulated event conditions are presented in Fig.~\ref{fig:ObjectPose}.
This figure presents the 6-DoF pose tracking results of objects in simulated event data from multiple viewpoints, where the green, red, and blue axes indicate the predicted 3D rotation and translation.
Meanwhile, the experimental results reported in Table~\ref{tab:Simulationcomparison}, compared to the state-of-the-art method, further validate the effectiveness of our proposed approach.
The results indicate that our method achieves robust 6-DoF pose tracking, maintaining stable and reliable object tracking performance even in complex and dynamic environments.




\subsection{Real Event Experiment} \label{sec:RealEventExperiment}
\begin{table}[t]
    \centering
    \renewcommand{\arraystretch}{1.1} 
    \setlength{\tabcolsep}{5pt} 
    \caption{Comparison of relative rotation error and relative translation error on the self-collected real event datasets [$R_{rel}$: deg/s, $T_{rel}$: cm/s].}
    \begin{tabular}{c|cc|cc|cc}
        \hline
        Method  & \multicolumn{2}{c|}{Condition} & \multicolumn{2}{c|}{EDOPT\cite{glover2024edopt}}  & \multicolumn{2}{c}{Ours} \\
        \hline
        Model & Environment & Speed & $R_{rel}$ & $T_{rel}$  & $R_{rel}$ & $T_{rel}$ \\
        \hline
        Part2 & Light & Slow & 1.21 & \textbf{4.17}   & \textbf{0.51} & 4.21 \\
        Part2 & Dark & Slow & 1.77 & 4.99  & \textbf{1.03} & \textbf{4.44} \\
        Part2 & Light & Fast & 0.81 & 7.21  & \textbf{0.44} & \textbf{5.06} \\
        Part2 & Dark & Fast & 1.66 & 6.98  & \textbf{0.49}  & \textbf{5.05} \\
        Part5 & Light & Slow & 7.32  & 5.66 & \textbf{0.75} & \textbf{3.32} \\
        Part5 & Dark & Slow & 5.11 & 6.09 & \textbf{0.54} & \textbf{4.76} \\
        Part5 & Light & Fast & 7.02 & 9.11  & \textbf{1.41} & \textbf{3.40} \\
            Part5 & Dark & Fast & 6.89 & 8.94  & \textbf{0.46} & \textbf{5.94} \\
        \hline
    \end{tabular}
    \vspace{-2mm}
    \label{tab:comparison}
\end{table}

In this section, we conduct practical experimental verification utilizing self-collected real event datasets. The overall experimental scene layout is illustrated in Fig. \ref{fig:SceneLayout}.
In the experimental setup, we use the Prophesee EVK4 event camera (1280 × 720 pixels) to capture event data from 6-DoF object pose changes, while the NOKOV motion capture system is used to acquire the corresponding 6-DoF ground truth pose. 


Both devices have been pre-calibrated to ensure spatial and temporal alignment. 
Subsequently, we select representative target objects with linear contour attributes and curved-surface contour attributes to conduct pose tracking experiments, followed by a systematic comparative analysis of the results.
The event camera is placed at a fixed distance from the objects, and we move the objects \redcolor{by hand} while the camera captures their motion and records the corresponding events. 
The 3D models of these objects are known in advance. Markers are attached to these objects to obtain the ground truth of their poses and trajectories using the NOKOV motion capture system.


In order to demonstrate the performance of our method in real-world scenarios, we conduct multiple sets of experiments, and the results are shown in Table \ref{tab:comparison}. 
The object tracking results using the event camera are shown in Fig. \ref{fig:RealPose}. 
The results indicate that the proposed method can achieve robust pose tracking in high-speed motion scenarios.


\section{CONCLUSIONS}
This paper proposes a keypoint-based approach for dynamic object 6-DoF pose tracking using event camera.  
The proposed method achieves stable 6-DoF pose tracking for a wide range of objects, including those with curved geometries, without requiring a predefined initialization.
We present a lightweight network for efficient event-based keypoints detection and propose a density-based tracking algorithm with an extended Kalman filter to achieve robust 6-DoF pose tracking under highly dynamic conditions.
Extensive experiments demonstrate that the proposed method outperforms state-of-the-art approaches on both synthetic and real-world datasets.
However, the proposed method requires known CAD models for object pose tracking. 
Future work will focus on developing approaches that enable pose tracking for a wider range of objects without relying on CAD models.


\bibliographystyle{IEEEtran}
\normalem
\bibliography{ref}

\end{document}